\documentclass{article}
\PassOptionsToPackage{table}{xcolor}
\usepackage{iclr2027_conference,times}

\usepackage{amsmath,amsfonts,bm}

\def\eqref#1{equation~\ref{#1}}

\def\1{\bm{1}}

\DeclareMathAlphabet{\mathsfit}{\encodingdefault}{\sfdefault}{m}{sl}
\SetMathAlphabet{\mathsfit}{bold}{\encodingdefault}{\sfdefault}{bx}{n}

\usepackage{url}
\usepackage{graphicx}
\usepackage{amsmath,amssymb}
\usepackage{booktabs}
\usepackage{xcolor}
\usepackage{placeins}
\usepackage{float}
\usepackage{wrapfig}
\usepackage{todonotes}
\usepackage{needspace}
\usepackage{listings}
\usepackage{enumitem}
\usepackage{multirow}
\usepackage{hyperref}
\usepackage{xcolor}
\newcommand{\std}[1]{{\color{gray}\scriptstyle\,\pm\,#1}}
\definecolor{darkorange}{HTML}{D04A16}
\graphicspath{{figs/}}


\definecolor{commentgray}{gray}{0.42}
\definecolor{numbergray}{gray}{0.55}
\lstdefinestyle{paperalgorithm}{
  language=Python,
  basicstyle=\small\ttfamily,
  keywordstyle=\color{black},
  commentstyle=\color{commentgray}\itshape,
  numbers=left,
  numberstyle=\scriptsize\color{numbergray},
  numbersep=1.1em,
  stepnumber=1,
  numberblanklines=false,
  frame=none,
  xleftmargin=2.5em,
  xrightmargin=0.5em,
  columns=fullflexible,
  keepspaces=true,
  showstringspaces=false,
  breaklines=true,
  aboveskip=0.65em,
  belowskip=0.65em
}
\newcommand{\ours}{\textcolor{darkorange}{\textbf{P-BFM}}}

\title{\textbf{\ours{}}: Perceptive Behavioral Foundation Models for Humanoid Control via Unsupervised Reinforcement Learning}

\usepackage{authblk}

\renewcommand{\Affilfont}{\normalfont\small}

\makeatletter
\renewcommand{\AB@affilsepx}{\quad\protect\Affilfont}
\makeatother
\author[1,3]{Xue~Wang\textsuperscript{*}}
\author[2]{Yitang~Li\textsuperscript{*}}
\author[3]{Yiming~Zhang\textsuperscript{*}}
\author[1]{Siqin~Wang}
\author[3,4]{Junhan~Sun}
\author[3,5]{Zhihao~Liu}
\author[3,4]{Yaojie~Shen}
\author[3]{Jingbo~Yuan}
\author[2]{Hao~Zhao}
\author[1]{Qingdu~Li\textsuperscript{\textdagger}}
\affil[1]{University of Shanghai for Science and Technology}
\affil[2]{Tsinghua University}
\affil[3]{RoboPartyLab}
\affil[4]{Zhejiang~University}
\affil[5]{Harbin Institute of Technology}
\newcommand{\authornotes}{\textsuperscript{*} Equal contribution.\qquad\textsuperscript{\textdagger} Corresponding author.}
\date{\authornotes}
\makeatletter
\@ifundefined{lx@add@date}{%
  \g@addto@macro\@maketitle{\vspace{-3\baselineskip}{\centering\small\authornotes\par}}%
}{}
\makeatother

\iclrfinalcopy
\hypersetup{
  pdftitle={P-BFM: Perceptive Behavioral Foundation Models for Humanoid Control via Unsupervised Reinforcement Learning},
  pdfauthor={Xue Wang, Yitang Li, Yiming Zhang, Siqin Wang, Junhan Sun, Zhihao Liu, Yaojie Shen, Jingbo Yuan, Hao Zhao, Qingdu Li}
}

\begin{document}
\maketitle
\fancyhead{}
\renewcommand{\headrulewidth}{0pt}

\begin{figure}[h]
    \centering
    \includegraphics[width=\linewidth]{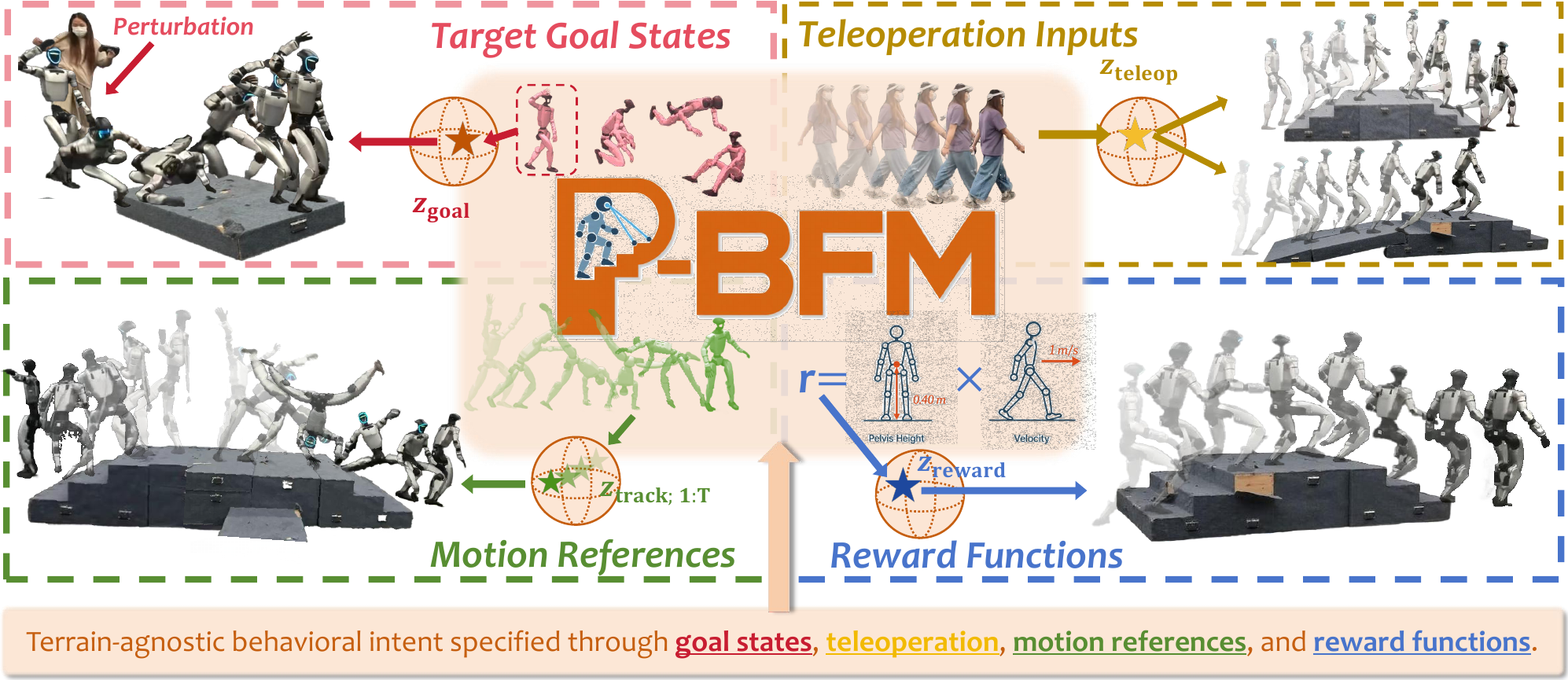}
    \vspace{-5mm}
\caption{Unified intent representation in \ours{}. Motion references, teleoperation inputs, target states, and reward functions are mapped into a shared terrain-agnostic intent space. A single perceptive policy realizes the same intent through distinct behaviors adapted to different terrains. Videos:
\url{https://roboparty.github.io/P-BFM/}}
    \label{fig:head}
\end{figure}

\begin{abstract}
Behavioral foundation models (BFMs) provide humanoids with reusable repertoires of whole-body behaviors; in real-world humanoid control, the same behavioral intent may require fundamentally different motions under different terrain constraints. Covering such variation with terrain-matched demonstrations requires enumerating a combinatorial space of behaviors and terrain geometries. We introduce \ours{}, a perceptive behavioral foundation model that factorizes humanoid control into terrain-invariant intent and terrain-conditioned realization. A forward–backward representation maps motion references, target states, and reward functions into a shared intent space without terrain information, while a perceptive policy learns how to realize each intent under the surrounding terrain geometry. We further identify a complementary limitation of purely latent control: behavioral latents can preserve local motion patterns while losing precise global progress. We therefore introduce Explicit Root Instructions (ERI), which separates global transport (e.g., heading and planar velocity) from behavioral intent, while leaving local whole-body adaptation to the perceptive policy. This decomposition leads to a characteristic capability absent from conventional BFMs: one fixed intent induces different, terrain-appropriate behaviors, including changes in step height, stride, posture, foothold placement, and contacts, without terrain-specific demonstrations. Across challenging terrains, \ours{} improves motion fidelity, root-control precision, and perturbation recovery over strong baselines. A single frozen policy further supports motion tracking, goal reaching, reward inference, teleoperation, and terrain-aware recovery, and transfers zero-shot to a Unitree G1 using onboard depth sensing.
\end{abstract}
\vspace{-2mm}

\section{Introduction}

\vspace{-2mm}


Humanoid robots are designed to operate in human-centric environments, with versatile embodiments suited to diverse physical tasks. Realizing this potential requires a reusable skill space and a unified controller capable of executing a broad range of skills. Recent advances in Behavior Foundation Models (BFMs) have made substantial progress in this direction, enabling a single controller to execute diverse behaviors through a shared skill space learned from large-scale Mocap datasets.

However, for humanoid robots operating in the real world, generality should also encompass perceiving and adapting to their surroundings. Many tracking-based BFMs rely heavily on demonstrations, and extending them to perceptive control substantially expands the data space to combinations of terrain geometry and motion. Covering this space with paired demonstrations is difficult, limiting generalization to unseen combinations. Prior approaches~\cite{pmt} address this bottleneck through terrain-aware motion augmentation, but the underlying optimization can yield suboptimal references, particularly for behaviors involving rich contact with the ground.

To address this limitation, we propose a \textcolor{darkorange}{\textbf{Perceptive Behavior Foundation Model}} (\ours{}), which learns through autonomous interaction with complex terrains rather than relying solely on terrain-matched motion imitation. We build on forward--backward (FB) representations, an unsupervised RL (Reinforcement Learning) framework that jointly learns behavioral representations and a latent-conditioned policy, allowing task-level reward functions to be mapped to latent variables for policy execution. These rewards map robot states to scalar values and specify behavioral intents, such as raising an arm or lying down. This exploration-driven framework enables the robot to acquire diverse behaviors without requiring demonstrations for terrain--motion combination.


Within this framework, the reward-specified intent should remain consistent across terrains, while its execution adapts to the terrain geometry. We therefore separate intent encoding from perception-driven execution. The backward representation encodes body-state features without terrain inputs, allowing the same reward specification to yield the same latent intent across terrains. The policy uses depth observations to realize intents specified by flat-ground rewards across diverse terrains. However, weak root-motion supervision in unsupervised learning can lead to heading drift or stepping in place before obstacles. We therefore introduce ERI (Explicit Root Instructions), which adds target heading and base-velocity instructions to improve root tracking while preserving terrain adaptation.

Experiments in simulation and the real world demonstrate that \ours{} outperforms all evaluated baselines in precision and robustness across terrains, supporting motion tracking, goal reaching, reward inference, fall recovery, and teleoperation within a shared perceptive policy. By realizing the same intent through terrain-adaptive behaviors without terrain-specific demonstrations, \ours{} enables humanoid robots to execute a shared repertoire of skills across different environments.

Our contributions are summarized as follows:

\begin{itemize}[leftmargin=10pt]
    \item We propose \ours{}, which uses unsupervised RL and FB representations to unify motions, goal states, and rewards in a terrain-agnostic intent space, enabling perceptive control across terrains without terrain-specific demonstrations.

    \item We introduce ERI, which complements unsupervised behavior learning with explicit heading and base-velocity supervision, enabling precise root control during terrain-adaptive execution.

    \item We demonstrate superior precision and robustness over existing baselines through extensive experiments and ablations, and validate diverse terrain-adaptive behaviors on a physical humanoid.
\end{itemize}

\section{Related Work}
\vspace{-1mm}

\subsection{Behavioral Foundation Models and Humanoid Whole-Body Control}
\vspace{-1mm}


Learning from human motion offers a practical foundation for expressive and diverse whole-body humanoid control. BeyondMimic~\cite{beyondmimic} combines robust motion tracking with guided diffusion to synthesize and compose motion primitives for downstream tasks. Similarly, NVIDIA's SONIC~\cite{sonic} scales motion tracking with large-scale human motion data to learn a unified whole-body controller for diverse humanoid behaviors. Complementarily, unsupervised RL explores behaviors without tracking specific motion references, aiming to learn reward-inferable representations through self-exploration. Recent methods based on forward--backward representations~\cite{touati2021learning}, including Meta Motivo~\cite{metamotivo}, BFM-Zero~\cite{bfmzero}, and UFO~\cite{ufo}, organize behaviors in a shared latent space where reward functions specify behavioral latents. They demonstrate diverse whole-body behaviors, perturbation recovery, and flexible task specification, but lack environmental perception for terrain-aware control. Their self-exploration capacity nevertheless offers a promising foundation for terrain adaptation without terrain-specific demonstrations. Building on this, \ours{} integrates depth perception with latent-conditioned control and learns through interaction with diverse terrains, allowing the same intent to be realized through terrain-adaptive behaviors.


\vspace{-1mm}
\subsection{Perceptive Locomotion and Terrain Adaptation}
\vspace{-1mm}

Perceptive humanoid control uses environmental observations to anticipate obstacles and adapt movement to terrain geometry~\cite{pim,ame,hikinginthewild,php,deepwholebodyparkour,pmt,ssr,duan2024,song2025,dpl,zhuang2024,pilot,nowyouseethat,more}. PIM~\cite{pim}, AME~\cite{ame}, Hiking in the Wild~\cite{hikinginthewild} and SSR~\cite{ssr} integrate terrain maps or depth observations for robust locomotion. Data-driven approaches, such as Perceptive Humanoid Parkour~\cite{php} and Deep Whole-body Parkour~\cite{deepwholebodyparkour}, enable dynamic whole-body skills but still depend on sufficient motion coverage and terrain-compatible demonstrations.
PMT~\cite{pmt} reduces this dependence through offline reference optimization and teacher--student training, but separating reference synthesis from policy learning can introduce discrepancies and complicate extension to contact-rich behaviors such as getting up.
In contrast, \ours{} learns terrain adaptation through unsupervised RL and autonomous interaction, without terrain-specific demonstrations or reference synthesis.
A shared policy uses latent intents to specify behavior and depth observations to adapt execution.



\vspace{-1mm}
\section{Preliminary}
\vspace{-2mm}
\subsection{Problem Formulation}
\vspace{-2mm}

We formulate perceptive humanoid control as a partially observable
Markov decision process
$\mathcal{M}=(\mathcal{S},\mathcal{O},\mathcal{A},P,\Omega,\gamma)$,
where $\mathcal{S}$, $\mathcal{O}$, and $\mathcal{A}$ denote the state,
observation, and action spaces, $P$ denotes transition dynamics,
$\Omega$ is the observation model, and $\gamma\in[0,1)$ is the
discount factor. The policy's observations include $o_t^p$ and
$h_t^v$, where $o_t^p$ contains proprioceptive information, previous
actions, and their history, and $h_t^v\in\mathbb{R}^{256}$ is the
temporally aggregated visual feature from 8 depth frames,
each represented by a 256-dimensional vector.
The action $a_t\in\mathbb{R}^{29}$ specifies joint targets executed
by a low-level PD controller. The policy receives $o_t^p$, including a four-frame history of
proprioception and previous actions, visual features $h_t^v$,
an estimated root velocity $\hat{\mathbf v}_t^{\mathrm{yaw}}$,
a latent $z_t$, and global heading and planar-velocity instructions
$c_{h,t}$ and $c_{v,t}$, yielding
$a_t\sim\pi_\theta(
\cdot\mid o_t^p,h_t^v,\hat{\mathbf v}_t^{\mathrm{yaw}},
z_t,c_{h,t},c_{v,t})$.
The latent $z_t$ specifies the desired behavior, visual features
guide terrain adaptation, and global instructions specify heading
and velocity requirements.

Our goal is to learn a shared latent-conditioned policy that
executes the behavior specified by $z_t$ across diverse terrains
while tracking global instructions, without task-specific policy
optimization. Training may use privileged information, including
true velocity and terrain information, whereas deployment relies
only on available sensor observations, the latent $z_t$, and
global instructions.

\vspace{-1mm}
\subsection{Preliminaries on Forward--Backward Representations}
\vspace{-1mm}

Forward--backward (FB) representations~\citep{touati2021learning} provide a reward-free framework for learning a family of latent-conditioned policies. For clarity, we introduce the standard formulation with fully observed states. Let $\rho$ be a reference state distribution. FB learns a forward representation $F:\mathcal{S}\times\mathcal{A}\times\mathbb{R}^{d}\rightarrow\mathbb{R}^{d}$ and a backward representation $B:\mathcal{S}\rightarrow\mathbb{R}^{d}$, where each latent $z\in\mathbb{R}^{d}$ indexes a policy $\pi_z$.

The discounted successor-state measure of $\pi_z$ is $M^{\pi_z}(ds'\mid s,a)=\sum_{k=0}^{\infty}\gamma^k\Pr(s_{t+k+1}\in ds'\mid s_t=s,a_t=a,\pi_z)$. FB approximates this measure with the low-rank factorization
\begin{equation}
M^{\pi_z}(ds'\mid s,a)
\approx F(s,a,z)^\top B(s')\,\rho(ds').
\label{eq:fb_factorization}
\end{equation}
For a downstream state reward $\tilde r$, the corresponding latent is $z_{\tilde r}=\mathbb{E}_{s\sim\rho}[B(s)\tilde r(s)]\approx \frac{1}{N}\sum_{i=1}^{N}B(s_i)\tilde r(s_i)$, with $s_i\sim\rho$. This gives $Q_{\tilde r}^{\pi_{z_{\tilde r}}}(s,a)\approx F(s,a,z_{\tilde r})^\top z_{\tilde r}$, so the policy can favor actions with high inner-product values. In \ours{}, rewards are defined on body-state features as $\tilde r(s)=r(\chi(s))$. This lets the reward-to-latent mapping specify behavioral intent, while the perception-conditioned policy handles terrain-dependent execution.

\vspace{-2mm}
\section{Method}
\vspace{-2mm}
\begin{figure}[t]
    \centering
    \includegraphics[width=0.9\linewidth]{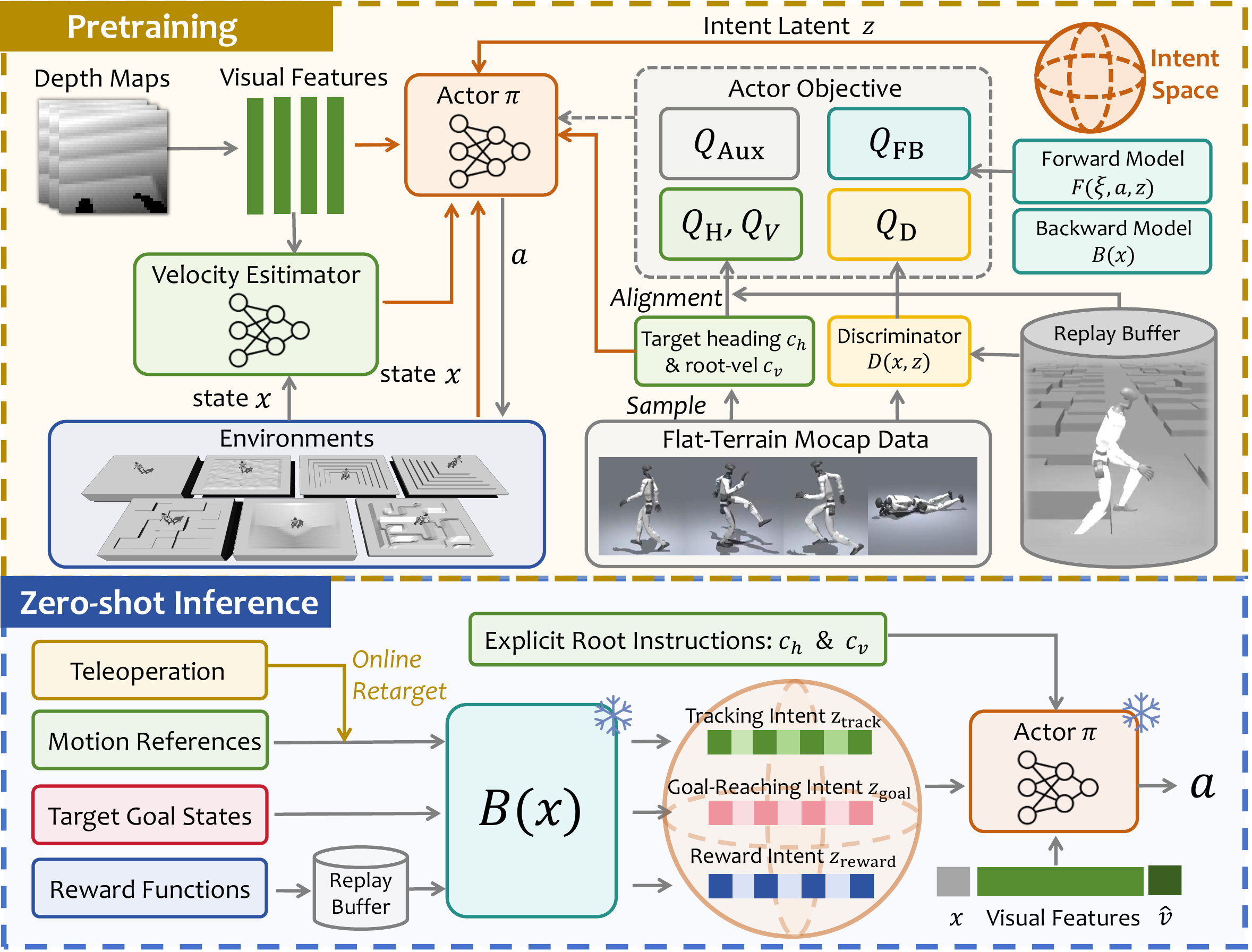}
\caption{
Overview of \ours{}. During pretraining, we learn a terrain-agnostic intent space through FB learning and a perceptive actor that interprets the intent latent into terrain-adaptive behavior. At inference, different task specifications are encoded into latents, and the same actor executes them across terrains without task-specific training. \textcolor{darkorange}{Orange arrows} denote inputs to the actor.
}

    \label{fig:overview}
\end{figure}

Our goal is to learn a single low-level humanoid controller that can realize diverse intents while adapting to complex terrain geometry. As shown in Fig.~\ref{fig:overview}, \ours{} consists of a terrain-agnostic intent space and a terrain-aware perceptive actor: the intent space specifies what behavior to perform, and the actor determines how to execute it with perception.

During pretraining, we encode depth observations into visual features for terrain perception (Sec.~\ref{sec:perception}), learn the intent space with FB representation, and train the actor with discriminator and auxiliary critics (Sec.~\ref{sec:method_training}). The FB objective enables zero-shot inference, the discriminator improves motion quality, and auxiliary critics facilitate sim-to-real transfer. To address the weak global-motion supervision in unsupervised RL, we introduce ERI (Explicit Root Instructions), which combines root heading and base-velocity instructions with intent latents to preserve commanded progress during terrain-adaptive execution (Sec.~\ref{sec:gili}). At inference, different task specifications are mapped to corresponding intent latents, including tracking intent, goal-reaching intent, and reward intent (Sec.~\ref{sec:method_inference}). The same perceptive actor then generates terrain-adaptive actions conditioned on the inferred latent, without task-specific policy optimization.

\vspace{-1mm}
\subsection{Perception Integration}
\label{sec:perception}
\vspace{-1mm}



To execute the same intent across terrains, the actor requires deployable terrain perception. Following the temporal aggregation and preprocessing design of Hiking in the Wild~\cite{hikinginthewild}, we maintain a 37-frame depth buffer, from which we sample every 5 frames to obtain $K=8$ observations with ages $(35,30,25,20,15,10,5,0)$, while applying a per-episode temporal offset of 0--1 frames to model sensing latency. Each $36\times64$ depth frame is cropped to $18\times32$, filtered with a $3\times3$ Gaussian kernel $(\sigma=1)$, clipped to $[0,2.5]$ meters, normalized. The resulting sequence is encoded by a shared CNN and aggregated with a GRU into $h_t^v$, which is used by the policy and velocity estimator to adapt execution to surrounding geometry.

Let $o_t^p$ denote proprioceptive inputs, including joint and inertial measurements, previous actions, and proprioceptive history. The estimator $\hat{\mathbf v}_t^{\mathrm{yaw}}=E_\eta(o_t^p,h_t^v)$ predicts the three-dimensional root linear velocity in a yaw-aligned frame. It receives no intent latent, global instructions, or true velocity, making it deployable. The actor is conditioned as
\begin{equation}
a_t\sim\pi_\theta\!\left(
\cdot\mid o_t^p,h_t^v,
\operatorname{sg}(\hat{\mathbf v}_t^{\mathrm{yaw}}), z_t,c_{h,t},c_{v,t}
\right),
\end{equation}
where $\operatorname{sg}$ stops gradients, $z_t$ specifies the intent, and $(c_{h,t},c_{v,t})$ are the global instructions introduced in Sec.~\ref{sec:gili}. Training may use privileged terrain and true velocity for critics or auxiliary losses, while deployment uses only onboard observations.

\vspace{-1mm}
\subsection{ERI: Explicit Root Instructions in Unsupervised RL}
\vspace{-1mm}
\label{sec:gili}
Although the intent is inferred from body-state features that include root states, unsupervised RL provides weak supervision for precise root-motion tracking. The policy may therefore preserve local patterns while drifting in heading or losing forward progress on challenging terrains. We address this with ERI (Explicit Root Instructions), which augments the intent-conditioned policy with target heading and base-velocity instructions, providing direct root-motion supervision while preserving perception-driven terrain adaptation.

For tracking intent, $z_t$ encodes the behavior from a short reference window, while ERI provides explicit target root heading and planar velocity from the same reference. Together, they encourage the robot to follow the reference motion instead of drifting or stepping in place before obstacles. Let $\psi_t,\psi_t^\star$ denote current and target headings, and $\mathbf v_{t,xy}^\star$ the target world-frame planar velocity. We encode the instructions as
\begin{equation}
\begin{aligned}
c_{h,t}
&=m_{h,t}
\begin{bmatrix}
1-\cos(\Delta\psi_t)\\
\sin(\Delta\psi_t)
\end{bmatrix},
\quad
\Delta\psi_t=\psi_t^\star-\psi_t,\\
c_{v,t}
&=
\begin{bmatrix}
m_{v,t}R(-\psi_t)\mathbf v_{t,xy}^\star\\
m_{v,t}
\end{bmatrix},
\end{aligned}
\end{equation}
where $R(-\psi_t)$ rotates the velocity instruction into the current heading frame, and $m_{h,t},m_{v,t}\in\{0,1\}$ mark valid targets. Zero velocity is treated as a valid stopping instruction; random or goal-conditioned latents do not automatically receive reference-velocity targets.

To train the policy to follow ERI, critics $Q_H(\xi,a,z)$ and $Q_V(\xi,a,z)$ learn from
\begin{equation}
\begin{aligned}
r_{H,t}
&=-m_{h,t}\left[
1-\cos(\psi_{t+1}^\star-\psi_{t+1})
\right],\\
r_{V,t}
&=-m_{v,t}\,\ell_{\mathrm{Huber}}\!\left(
\frac{
\|\mathbf v_{t+1,xy}^{\mathrm{sim}}
-\mathbf v_{t,xy}^\star\|_2
}{\sigma_v} \right).
\end{aligned}
\end{equation}
The heading reward uses the next reference heading, while the velocity reward compares post-action planar velocity with the current instruction. Since $Q_V$ constrains only planar velocity, the policy can still adapt vertical motion to the terrain. Here, $\ell_{\mathrm{Huber}}(u)=u^2/2$ for $0\leq u\leq1$ and $u-1/2$ otherwise, and $\sigma_v$ sets the velocity-error scale.

\vspace{-1mm}
\subsection{Pretraining of \ours{}}
\vspace{-1mm}
\label{sec:method_training}

Pretraining learns two coupled components: a terrain-agnostic intent space and a terrain-aware actor. We implement $B(s)=B_\phi(x)$, where $x=\chi(s)$ contains proprioceptive and privileged body-state information, but no depth, terrain maps, or instructions. In contrast, $F_\psi(\xi,a,z)$ receives history, terrain information, and instructions through $\xi$. The FB score is $Q_{\mathrm{FB}}(\xi,a,z)=F_\psi(\xi,a,z)^\top z$, thus the same target body state or reward is encoded into the same latent regardless of terrain, while $F_\psi$ and the actor handle terrain-dependent execution.

The policy is trained by online interaction with varied simulated terrains. Forty flat-ground LaFAN sequences provide behavior prompts and positives for the conditional discriminator $D_\omega(x,z)$, without terrain-specific motion synthesis. We train $F$ and $B$ with the FB Bellman objective and orthonormality regularization. The discriminator reward $Q_D$, physical auxiliary critic $Q_A$, and ERI critics $Q_H,Q_V$ encourage natural motion, physical regularity, and root tracking.

The actor maximizes the FB value and auxiliary critics:
\begin{equation}
\mathcal L_\pi
=-\overline Q_{\mathrm{FB}}
-\alpha\sum_{k\in\{D,A,H,V\}}\lambda_k\overline Q_k,
\qquad
\alpha=\operatorname{sg}\!\left(
\mathbb E[|Q_{\mathrm{FB}}|]
\right).
\end{equation}
Here, $\overline Q_k$ averages conservative critic estimates, and $H,V$ are used only for valid ERI targets.

We also train the deployable velocity estimator with simulator velocity:
\begin{equation}
\mathcal L_{\mathrm{est}}
=\frac{1}{3}\mathbb E\!\left[
\|\hat{\mathbf v}_t^{\mathrm{yaw}}
-\mathbf v_{t,\mathrm{sim}}^{\mathrm{yaw}}\|_2^2 \right].
\end{equation}
We minimize $\mathcal L_\pi+\lambda_{\mathrm{est}}\mathcal L_{\mathrm{est}}$. The estimation loss updates the estimator and depth encoder, while policy gradients are stopped through the estimated velocity.

\subsection{Zero-shot Inference of \ours{}}
\label{sec:method_inference}

With all parameters fixed, \ours{} supports three intent types: \emph{tracking intent}, \emph{goal-reaching intent}, and \emph{reward intent}, derived from motion references, target goal states, and reward functions, respectively. All intents are encoded through the frozen backward representation:
\begin{equation}
z=
\begin{cases}
\displaystyle
\mathcal P\!\left(
\frac{1}{L_t}
\sum_{k=0}^{L_t-1}B_\phi(x_{t+k}^{\mathrm{ref}})
\right),
& \text{tracking intent},\\[6pt]
\displaystyle
\mathcal P\bigl(B_\phi(x^\star)\bigr),
& \text{goal-reaching intent},\\[6pt]
\displaystyle
\mathcal P\!\left(
\frac{1}{N}\sum_{i=1}^{N}r(x_i)B_\phi(x_i)
\right),
& \text{reward intent}.
\end{cases}
\label{eq:intent_encoding}
\end{equation}
Here, $x_{t+k}^{\mathrm{ref}}$ denotes an aligned reference body state, $x^\star$ is a target body state, and $x_i\sim\mu$ are replay body-state features. For a reference sequence of length $T$, we use $L_t=\min(8,T-t)$. The reward $r(x)$ specifies the desired behavior, and $\mathcal P$ normalizes nonzero vectors to norm $\sqrt d$, matching the latent scale used in training. Teleoperation uses the same tracking interface with online retargeted body-state targets. At deployment, the policy executes its mean action and uses ERI heading and planar-velocity instructions whenever valid.
\vspace{-2mm}
\section{Experiments}
\vspace{-2mm}

We evaluate whether \ours{} can translate different forms of behavioral intent into robust terrain-adaptive control. Following Sec.~\ref{sec:method_inference}, we consider three intent types: \emph{goal-reaching intent}, \emph{tracking intent}, and \emph{reward intent}. These intents are derived from target goal states, motion references, and reward functions, respectively. They correspond to three main evaluation settings: goal reaching, motion tracking, and reward inference. We further include teleoperation to test whether the learned policy can follow online user instructions while adapting to real-world terrain.

Across these settings, our evaluation is organized around four questions:
\begin{itemize}[leftmargin=8pt]
    \item Does \ours{} outperform existing general-purpose humanoid controllers in motion tracking on challenging terrains?
    \item Can \ours{} generate terrain-adaptive behaviors from the same intent across diverse terrains?
    \item How do explicit root instructions improve control precision and terrain traversal?
    \item How does \ours{} perform on a real humanoid robot using onboard sensing?
\end{itemize}

\paragraph{Experimental Setup and Metrics.}
We train \ours{} using Unitree G1 in MJLab~\cite{mjlab} for 192 million environment steps on NVIDIA H200×8 GPUs in approximately 10 hours and perform zero-shot inference with all model parameters frozen. Optimization settings are provided in Appendix~\ref{app:optimization}.

We report joint-angle MAE \(E_{\mathrm{jpos}}\),
root horizontal-velocity error \(E_{\mathrm{vel}}^{\mathrm{root}}\),
root-heading error \(E_{\mathrm{head}}^{\mathrm{root}}\),
foot sliding (FS; mean displacement in cm at 50~Hz), and tracking recovery rate (RR).
Recovery requires heading error \(<30^\circ\) and horizontal-velocity error \(\leq0.5\,\mathrm{m/s}\) within 5~s after perturbation removal.Both conditions must hold continuously for at least 1~s and remain satisfied until the 10~s rollout ends. Trials remaining within tolerance also count as successful. We evaluate all methods on 7 terrain types: flat, rough, ascending stairs, descending stairs, platforms, inverted pyramid slopes, and box obstacles. Each type has 10 difficulty levels, with flat terrain unchanged across levels. Appendix~\ref{app:terrains} details the terrains; Appendices~\ref{app:continuous-metrics}, \ref{app:recovery}, and \ref{app:aggregation} define the metrics, recovery protocol, and aggregation.

    \vspace{-2mm}
\subsection{Comparison with representative humanoid control baselines.}
    \vspace{-1mm}
\label{sec:baseline_comparison}

To answer the first question, we compare \ours{} with general-purpose humanoid controllers UFO and PMT. UFO is an unsupervised RL framework based on FB representations, while PMT adapts generic reference motions to terrains through offline optimization and incorporates terrain perception into reference-motion tracking. Under matched robot models and terrain conditions, we evaluate tracking accuracy, motion fidelity, and recovery on challenging terrains.

Since PMT's full training dataset is not publicly available, its motion coverage cannot be verified. We therefore evaluate its released model on the only reference trajectory provided, while evaluating \ours{} and UFO on a broader set of 40 LaFAN sequences. Reference-motion coverage and trial counts are detailed in Appendix~\ref{app:motion-coverage}. As shown in Table~\ref{tab:controller_comparison}, despite this broader motion coverage, \ours{} achieves lower joint-position and heading errors, less foot sliding, and substantially higher recovery rates than the baselines. Although the differing evaluation sets preclude a strictly matched comparison, these results suggest higher motion style reproduction accuracy and robustness for \ours{}. We attribute these gains to end-to-end learning through autonomous exploration, which enables motion adaptation through direct interaction with diverse terrains rather than offline reference-motion optimization.

Under matched evaluation conditions on the unseen motion dataset, \ours{} consistently outperforms the baselines in motion fidelity, root-instruction tracking, and recovery, demonstrating stronger generalization beyond its training motions. The lower heading and velocity errors highlight the effectiveness of explicit root instructions, while the higher recovery rate suggests that unsupervised exploration helps the policy learn to recover from diverse off-nominal states.


\begin{table}[t]
\centering
\small
\caption{Comparison with existing perception-based general-purpose humanoid controllers on seen motions and a subset of 994 unseen SOMA motions.}
\vspace{2mm}
\label{tab:controller_comparison}
\renewcommand{\arraystretch}{1.15}
\setlength{\tabcolsep}{3.8pt}
\resizebox{\linewidth}{!}{
\begin{tabular}{lcccccccccc}
\toprule
\multirow{2}{*}{Method}
& \multicolumn{5}{c}{Seen motions}
& \multicolumn{5}{c}{Unseen motions} \\
\cmidrule(lr){2-6} \cmidrule(lr){7-11}
& $E_{\mathrm{jpos}}\;(\mathrm{rad})\downarrow$
& $E_{\mathrm{vel}}^{\mathrm{root}}\;(\mathrm{m/s})\downarrow$
& $E_{\mathrm{head}}^{\mathrm{root}}\;({}^{\circ})\downarrow$
& $\mathrm{FS}\;(\mathrm{cm})\downarrow$
& $\mathrm{RR}\;(\%)\uparrow$
& $E_{\mathrm{jpos}}\;(\mathrm{rad})\downarrow$
& $E_{\mathrm{vel}}^{\mathrm{root}}\;(\mathrm{m/s})\downarrow$
& $E_{\mathrm{head}}^{\mathrm{root}}\;({}^{\circ})\downarrow$
& $\mathrm{FS}\;(\mathrm{cm})\downarrow$
& $\mathrm{RR}\;(\%)\uparrow$ \\
\midrule
UFO
& $0.15\std{0.06}$ & $0.65\std{0.46}$
& $49.6\std{29.3}$ & $0.092\std{0.050}$ & $3.4$
& $0.16\std{0.06}$ & $0.37\std{0.27}$
& $28.9\std{23.9}$ & $0.066\std{0.047}$ & $16.6$ \\

PMT
& $0.21\std{0.04}$ & $\mathbf{0.20}\std{0.08}$
& $8.8\std{7.7}$ & $0.078\std{0.029}$ & $2.4$
& $0.46\std{0.09}$ & $0.86\std{0.27}$
& $84.4\std{22.5}$ & $0.205\std{0.103}$ & $0.5$ \\

\ours{}
& $\mathbf{0.11}\std{0.04}$ & ${0.25}\std{0.12}$
& $\mathbf{6.9}\std{5.4}$ & $\mathbf{0.043}\std{0.026}$
& $\mathbf{41.2}$
& $\mathbf{0.13}\std{0.06}$ & $\mathbf{0.21}\std{0.12}$
& $\mathbf{6.7}\std{4.7}$ & $\mathbf{0.026}\std{0.016}$
& $\mathbf{61.4}$ \\
\bottomrule
\end{tabular}}
\vspace{1mm}
\begin{minipage}{\linewidth}
\footnotesize
\textsuperscript{*}For seen-motion evaluation, PMT is evaluated on its only released reference trajectory, whereas UFO and \ours{} are evaluated on 40 LaFAN sequences.
\end{minipage}
\end{table}

    \vspace{-1mm}
\subsection{Terrain-Adaptive Behavior from a Single Intent}
    \vspace{-1mm}

\begin{figure}[h]
    \centering
    \includegraphics[width=\linewidth]{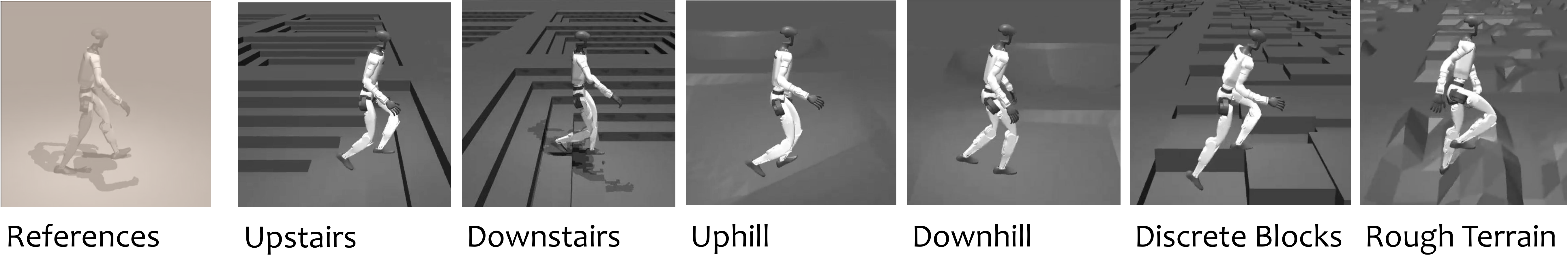}
    \vspace{-3mm}
    \caption{{Terrain-adaptive behavior from a single forward-walking intent. Given the same reference-derived intent, \ours{} produces distinct locomotion patterns across terrains, including stair ascent/descent, slope traversal, discrete foothold selection, and rough-terrain stepping.}}
    \label{fig:terrain_adapt}
\end{figure}

We examine whether \ours{} can realize the same reward-inferred intent through terrain-dependent motion patterns. Using forward walking as an example, we define a reward function for flat-ground walking and infer the corresponding latent intent through reward inference. This single latent is then evaluated across multiple terrains. As shown in Fig.~\ref{fig:terrain_adapt}, the policy does not simply reproduce a fixed walking pattern. Instead, it autonomously changes step height, stride length, body posture, and foothold selection according to the perceived terrain geometry. It is worth noting that no terrain-specific motion demonstrations are provided during training. The adaptive behaviors emerge from autonomous exploration in complex environments rather than from terrain-matched reference motions, demonstrating that \ours{} can discover distinct ways to realize the same reward-inferred intent under different terrain constraints.


\subsection{Ablation Study}
\label{sec:ablation}

We evaluate ERI by separately removing heading and velocity instructions from \ours{}. Table~\ref{tab:ablation} shows that both instructions substantially improve their respective tracking metrics on seen and unseen motions, with relatively modest differences in the remaining tracking and foot-skating metrics.

\begin{figure}[h]
    \centering
    \includegraphics[width=\linewidth]{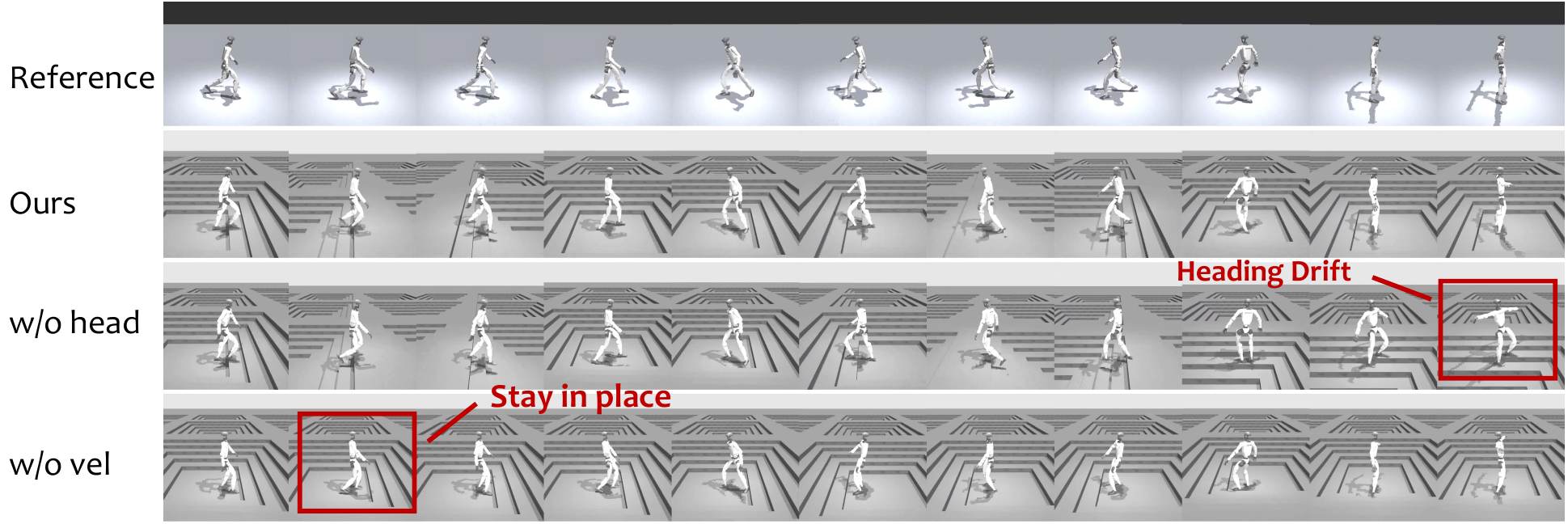}
    \vspace{-3mm}
    \caption{{Qualitative ablation. Without heading instructions, the robot exhibits heading drift; without velocity instructions, it steps in place. \ours{} maintains the intended heading and progress.}}
    \label{fig:ablation}
\end{figure}

Fig.~\ref{fig:ablation} illustrates the role of each instruction: without heading instructions, the robot gradually drifts from the intended direction; without velocity instructions, it reproduces local stepping motions but makes little forward progress when facing challenging terrain. In contrast, \ours{} preserves the reference behavior while maintaining the intended heading and forward progress, demonstrating the importance of both instructions for effective terrain traversal. ERI thus combines unsupervised behavioral exploration with supervised instruction tracking, preserving behavioral diversity while enforcing explicit root-motion requirements.

\begin{table*}[t]
\centering
\caption{Ablation study of global instructions on seen and unseen motions. ``w/o head'' removes the global heading instruction, while ``w/o vel'' removes the global base-velocity instruction.}
\label{tab:ablation}
\resizebox{\textwidth}{!}{
\begin{tabular}{lccccc@{\hspace{16pt}}ccccc}
\toprule
\multirow{2}{*}{Method}
& \multicolumn{5}{c}{Seen motions}
& \multicolumn{5}{c}{Unseen motions} \\
\cmidrule(lr){2-6}\cmidrule(lr){7-11}
& $E_{\mathrm{jpos}}\;(\mathrm{rad})\downarrow$
& $E_{\mathrm{vel}}^{\mathrm{root}}\;(\mathrm{m/s})\downarrow$
& $E_{\mathrm{head}}^{\mathrm{root}}\;({}^{\circ})\downarrow$
& $\mathrm{FS}\;(\mathrm{cm})\downarrow$
& $\mathrm{RR}\;(\%)\uparrow$
& $E_{\mathrm{jpos}}\;(\mathrm{rad})\downarrow$
& $E_{\mathrm{vel}}^{\mathrm{root}}\;(\mathrm{m/s})\downarrow$
& $E_{\mathrm{head}}^{\mathrm{root}}\;({}^{\circ})\downarrow$
& $\mathrm{FS}\;(\mathrm{cm})\downarrow$
& $\mathrm{RR}\;(\%)\uparrow$ \\
\midrule
w/o head
& $0.12\std{0.04}$ & $0.28\std{0.14}$
& $19.3\std{14.2}$ & $0.045\std{0.026}$ & $20.7$
& $\mathbf{0.13}\std{0.06}$ & $0.22\std{0.12}$
& $14.6\std{14.3}$ & $0.026\std{0.018}$ & $39.5$ \\

w/o vel
& $0.12\std{0.04}$ & $0.41\std{0.26}$
& $7.7\std{6.9}$ & $\mathbf{0.032}\std{0.018}$ & $22.4$
& $\mathbf{0.13}\std{0.06}$ & $0.32\std{0.19}$
& $\mathbf{6.3}\std{4.9}$ & $\mathbf{0.020}\std{0.012}$ & $42.9$ \\

\ours{}
& $\mathbf{0.11}\std{0.04}$ & $\mathbf{0.25}\std{0.12}$
& $\mathbf{6.9}\std{5.4}$ & $0.043\std{0.026}$
& $\mathbf{41.2}$
& $\mathbf{0.13}\std{0.06}$ & $\mathbf{0.21}\std{0.12}$
& $6.7\std{4.7}$ & $0.026\std{0.016}$
& $\mathbf{61.4}$ \\
\bottomrule
\end{tabular}}
\end{table*}

    \vspace{-1mm}
\subsection{Real-world Validation}
\label{sec:sim_to_real}

We deploy \ours{} on a Unitree G1 humanoid equipped with a RealSense D435i depth camera to examine its transfer from simulation to the real world. The experiments cover teleoperation, motion tracking, goal reaching, reward inference, and terrain-adaptive recovery from external perturbations. Camera settings and depth preprocessing are reported in Appendix~\ref{app:realworld-depth}.

\begin{figure}[h]
    \centering
    \includegraphics[width=\linewidth]{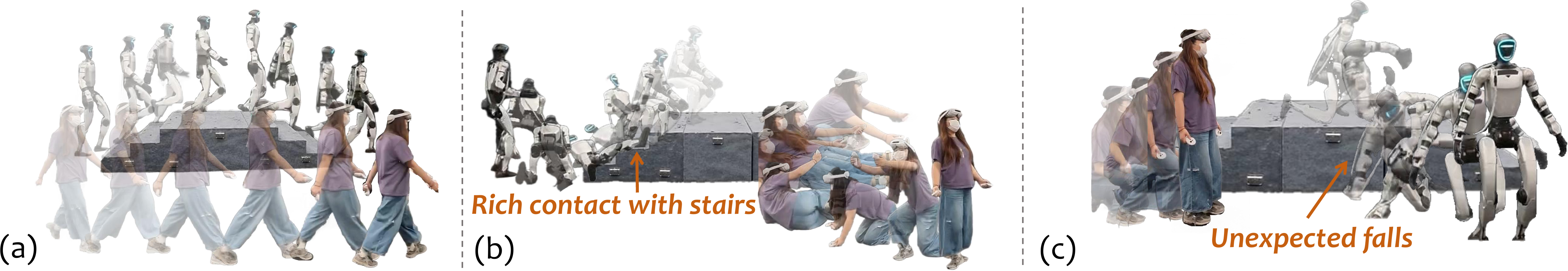}
        \vspace{-3mm}
    \caption{Real-world teleoperation with perception-driven terrain adaptation.}
    \label{fig:tele_real}
\end{figure}

\paragraph{Teleoperation.}
Fig.~\ref{fig:tele_real} shows real-world teleoperation with perception-driven terrain adaptation. In Fig.~\ref{fig:tele_real}(a), the robot follows teleoperation-derived walking intent and adapts it to stair traversal. Fig.~\ref{fig:tele_real}(b) shows contact-rich interaction with stairs, where the robot preserves the intended whole-body behavior despite frequent body-terrain contacts. Fig.~\ref{fig:tele_real}(c) further shows recovery from an unexpected fall, demonstrating robustness beyond nominal execution.

Also. as illustrated in the upper-left panel of Fig.~\ref{fig:head}, the same teleoperation input leads to different executions on stairs and slopes: higher foot lifting and discrete foothold placement on stairs, and smoother continuous contact adjustment on slopes. These terrain-dependent gait patterns preserve the intended behavior rather than literally reproducing the human motion.

\paragraph{Motion Tracking.}
As shown in Fig.~\ref{fig:head}(c), \ours{} tracks expressive whole-body motion references, including dynamic cartwheels and dances, on and around an elevated platform. Despite terrain constraints, the robot preserves the reference motion style while adapting foot placement, body posture, and contacts through perception. This demonstrates that tracking intent can be realized in terrain-aware ways within the same policy.

\begin{figure}[h]
    \centering
    \includegraphics[width=\linewidth]{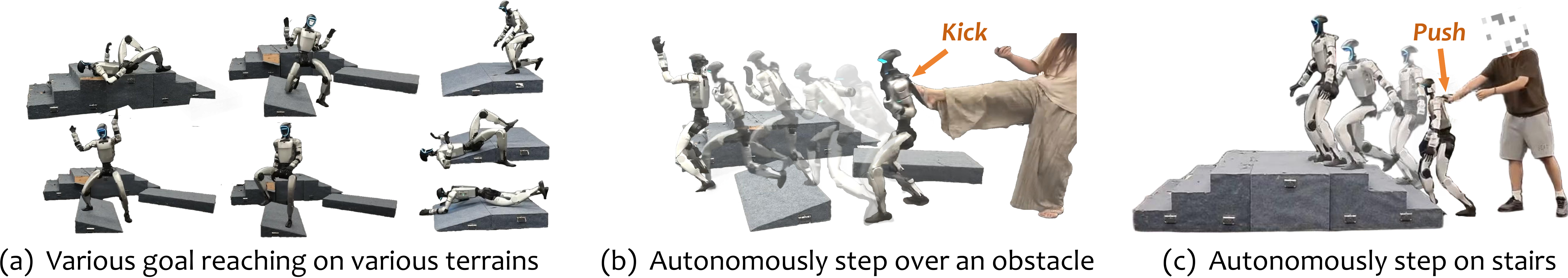}

        \vspace{-2mm}
        \caption{Real-world demonstrations of goal reaching and terrain-aware recovery.}
    \label{fig:goal}
\end{figure}

\paragraph{Goal Reaching.}
As shown in the upper-left part of Fig.~\ref{fig:head}, \ours{} supports any-to-any goal reaching and adaptive behavior across terrain variations. Fig.~\ref{fig:goal}(a) further presents diverse goal-reaching results on different terrains. Starting from different initial configurations, the robot reaches target poses such as sitting, lying down, standing, and arm-raising on stairs, slopes, gaps, and uneven platforms. These results show that goal-reaching intent is not tied to a specific terrain or initial state. During the transition, the robot adapts its contacts and body posture to the surrounding geometry, demonstrating terrain-aware goal reaching.

\paragraph{Reward Inference.}
In lower-right part of Fig.~\ref{fig:head}, the reward specifies forward velocity and pelvis-height preferences.
\ours{} infers the corresponding latent intent and executes it on stair-like terrain, maintaining forward progress while adapting its steps to height changes.
This shows that reward-derived intents can be grounded in perceptive terrain-adaptive control without additional planning, post-training or motion optimization.

\paragraph{Adaptive Behavior to Perturbation.}
Fig.~\ref{fig:goal}(b) and Fig.~\ref{fig:goal}(c) show terrain-aware recovery under external perturbations.
When kicked near an obstacle, the robot lifts its swing leg and steps over the block to recover balance.
When pushed near stairs, it autonomously places its feet on the stair surfaces and steps upward instead of colliding with the step edges.
These behaviors show that perception helps convert disturbances into feasible terrain-adaptive recovery motions.

    \vspace{-1mm}
    \section{Conclusion}
    \vspace{-1mm}
We introduced \ours{}, a Perceptive Behavior Foundation Model that unifies motion references, target states, and rewards in a terrain-agnostic intent space. A shared perceptive policy realizes these intents across terrains, while ERI complements unsupervised behavior learning with explicit root-motion supervision. Simulation and real-world experiments demonstrate motion tracking, goal reaching, reward inference, teleoperation, and terrain-adaptive recovery without terrain-specific demonstrations. However, behavioral generalization remains limited by the diversity of motion priors and terrains explored during training. Future work will investigate broader motion datasets and more challenging terrains, particularly those requiring complex whole-body contacts.

\section*{AI Use Statement}
We used generative AI tools for language editing and
translation, literature retrieval and reference
organization, and feedback on our methodology,
experimental protocols, and interpretation of results.
AI tools also assisted in drafting and revising selected
passages and preparing draft schematic figures.
The authors take full responsibility for the final
methodology, experiments, results, claims, text,
and figures presented in this work.

\bibliographystyle{iclr2027_conference}
\begingroup
\interlinepenalty=10000
\bibliography{refs_pbfm_core}
\endgroup

\clearpage
\appendix
\section{Additional Experimental Details}
\label{app:additional-details}
\subsection{Network Architectures}
\label{app:network-architectures}

The behavioral latent has dimension \(d=256\), and the actor
outputs 29 joint-action components.
The backward representation and behavioral latents are normalized
to Euclidean norm \(\sqrt{d}=16\).
Table~\ref{tab:network-inputs} summarizes the observation groups.

\begin{table}[H]
    \centering
    \small
    \caption{Observation groups used by the full model.
    Privileged inputs are used during training and are excluded
    from the deployed actor.}
    \label{tab:network-inputs}
    \begin{tabular}{l r}
        \toprule
        Observation group & Dimension \\
        \midrule
        Proprioceptive state & 64 \\
        Previous action & 29 \\
        Four-step proprioception--action history & 372 \\
        Heading instruction & 2 \\
        Horizontal-velocity instruction and validity flag & 3 \\
        Estimated current root velocity & 3 \\
        Privileged body state & 463 \\
        Privileged terrain features & 273 \\
        Ground-truth velocity supervision & 3 \\
        Aggregated visual feature & 256 \\
        \bottomrule
    \end{tabular}
\end{table}

The 64-dimensional proprioceptive state comprises 29 relative
joint positions, 29 joint velocities, projected gravity, and
root angular velocity.
Each historical observation contains these 64 features and
29 action components, giving \(4\times93=372\) history features.

\paragraph{Residual blocks.}
Let \(L_{p,q}\) denote an affine map from
\(\mathbb R^p\) to \(\mathbb R^q\).
We use pre-normalized projection and residual blocks:
\begin{align}
P_{p,q}(x)
&=
\operatorname{Mish}
\left(L_{p,q}(\operatorname{LN}_p(x))\right),\\
H_{p,q}(x)
&=
L_{p,q}(\operatorname{LN}_p(x)),\\
R_w(x)
&=
x+P_{w,w}(x).
\end{align}
Each residual block contains one affine layer.
The output projection \(H\) has no activation.

\paragraph{Forward representation and scalar critics.}
The forward representation \(F\) and scalar critics
\(Q_D,Q_A,Q_H,Q_V\) share the same input organization,
but have separate parameters.
Here, \(Q_A\) denotes the physical auxiliary critic,
also written as \(Q_{\mathrm{aux}}\).

Their training observation vector
\(o_c\in\mathbb R^{1209}\) concatenates proprioception,
privileged body state, previous action, history,
privileged terrain features, heading and velocity instructions,
and ground-truth velocity features.
Two branches encode \([o_c,z]\in\mathbb R^{1465}\) and
\([o_c,a]\in\mathbb R^{1238}\), respectively.
Each branch has the form
\begin{equation}
E_{p,w,k}
=
P_{w,w/2}\circ R_w^k\circ P_{p,w},
\end{equation}
where \(R_w^k\) denotes \(k\) separately parameterized
residual blocks.
The two branch outputs are concatenated and processed by
\(r\) additional residual blocks and an output projection
\(H_{w,m}\).

\begin{table}[H]
    \centering
    \small
    \caption{Forward and critic architectures.
    Each branch includes two affine projections in addition
    to its listed residual blocks.
    Each trunk includes an additional output projection.
    Each ensemble member is a complete, independently
    parameterized network.}
    \label{tab:critic-architectures}
    \begin{tabular}{l r r r r r}
        \toprule
        Network & Width \(w\) & Branch blocks \(k\)
        & Trunk blocks \(r\) & Output \(m\) & Members \\
        \midrule
        \(F\)   & 2048 & 4 & 6 & 256 & 2 \\
        \(Q_D\) & 1024 & 2 & 4 & 1   & 2 \\
        \(Q_A\) & 2048 & 2 & 4 & 1   & 2 \\
        \(Q_H\) & 1024 & 1 & 3 & 1   & 2 \\
        \(Q_V\) & 1024 & 1 & 3 & 1   & 2 \\
        \bottomrule
    \end{tabular}
\end{table}

\paragraph{Backward representation.}
The backward network receives the 527-dimensional concatenation
of proprioceptive and privileged body-state features.
Its architecture is
\begin{equation}
527
\xrightarrow{\mathrm{Linear}}
256
\xrightarrow{\mathrm{LN},\,\tanh}
256
\xrightarrow{\mathrm{Linear}}
256
\xrightarrow{\ell_2\text{ normalization}}
B(x),
\end{equation}
where the final normalization scales the output norm to 16.
Thus, \(B\) has one hidden layer and two affine layers.
It receives neither terrain observations nor global instructions.

\paragraph{Actor.}
The actor's low-dimensional observation contains 470 features.
Appending the detached three-dimensional velocity estimate
produces \(u_t\in\mathbb R^{473}\).
Two branches encode \(u_t\) and \([u_t,z_t]\) using
\[
473\rightarrow2048\rightarrow1024,
\qquad
729\rightarrow2048\rightarrow1024,
\]
with a projection block \(P\) at each arrow.
Their outputs are concatenated with the 256-dimensional visual
feature and fused by \(P_{2304,2048}\).
The policy trunk contains six \(R_{2048}\) blocks,
followed by \(H_{2048,29}\) and \(\tanh\).
Training uses action sampling with standard deviation \(0.05\);
evaluation uses the mean action.

\paragraph{Discriminator.}
The conditional discriminator receives
\([x,z]\in\mathbb R^{783}\) and has layer widths
\[
783\rightarrow1024\rightarrow1024\rightarrow1024\rightarrow1.
\]
The first hidden layer uses layer normalization and \(\tanh\);
the next two hidden layers use ReLU.
The output is a scalar logit, with a sigmoid applied when
a probability is required.

\paragraph{Observation normalization.}
Proprioception, privileged body state, previous action, history,
and privileged terrain features are standardized using running
statistics with momentum \(0.01\).
This normalization has no learned affine parameters.
Heading and velocity instructions bypass this standardization.
Depth normalization is described below.

\subsection{Depth Processing and Visual Encoding}
\label{app:depth-processing}

\paragraph{Simulated depth observations.}
The simulated sensor produces \(36\times64\) images
(height \(\times\) width), with horizontal and vertical
fields of view of \(89.51^\circ\) and \(58.29^\circ\).
Ray distances are converted to optical-axis depth using
the corresponding unit-ray direction.
The depth-processing profile maps invalid depth observations
to the maximum distance and clips depths to
\([0,2.5]\,\mathrm{m}\).

We retain rows \(18{:}36\) and columns \(16{:}48\), using
zero-based, end-exclusive indexing.
This produces an \(18\times32\) image without resizing.
The crop is smoothed using a \(3\times3\) Gaussian kernel
with standard deviation 1 pixel and reflection padding.
The filtered depth \(\bar d_t\) is clipped and quantized as
\begin{equation}
u_t
=
\operatorname{uint8}
\left[
\operatorname{round}
\left(
255\frac{\bar d_t}{2.5\,\mathrm{m}}
\right)
\right].
\end{equation}
The encoder converts the stored bytes to floating-point
values by dividing by 255.

\paragraph{Temporal sampling and reset.}
At the 50~Hz control rate, the visual input contains eight
frames sampled from a 37-frame buffer:
\begin{equation}
\mathcal D_t
=
\left[
u_{t-\delta-35},
u_{t-\delta-30},
\ldots,
u_{t-\delta-5},
u_{t-\delta}
\right],
\qquad
\delta\sim\operatorname{Uniform}\{0,1\}.
\end{equation}
Frames are ordered from oldest to newest.
Adjacent selected frames are separated by \(0.1\,\mathrm{s}\),
and the selected window spans \(0.7\,\mathrm{s}\).
The delay \(\delta\) is sampled at reset and remains fixed
within the episode.
On reset, the image buffer is filled with the first image
of the new episode, preventing visual history from carrying
over from the preceding episode.

\paragraph{Camera randomization.}
At each reset, camera translation is perturbed independently
within \(\pm0.01\,\mathrm{m}\) along each camera-link axis.
Roll, pitch, and yaw perturbations are sampled within
\(\pm2^\circ\), \(\pm10^\circ\), and \(\pm2^\circ\), respectively.
These extrinsic perturbations remain fixed within an episode.
The selected processing profile uses deterministic Gaussian
smoothing without additional stochastic image-intensity augmentation.

\paragraph{CNN and temporal aggregation.}
A shared CNN processes each depth frame independently.
Its architecture is given in Table~\ref{tab:depth-encoder}.
The eight resulting 256-dimensional frame features are passed
to a single-layer, unidirectional GRU with hidden dimension 256.
The final hidden state is the visual feature \(h_t^v\).
Each eight-frame window is encoded from a zero GRU initial state;
hidden states are not carried across policy evaluations.

\begin{table}[t]
    \centering
    \small
    \caption{Shared single-frame depth encoder.
    Shapes are channels \(\times\) height \(\times\) width
    before flattening.}
    \label{tab:depth-encoder}
    \begin{tabular}{l l l}
        \toprule
        Layer & Parameters & Output shape \\
        \midrule
        Input & Normalized depth & \(1\times18\times32\) \\
        Conv + ReLU & \(1\to16,\ k=3,\ s=1,\ p=1\)
                    & \(16\times18\times32\) \\
        Conv + ReLU & \(16\to32,\ k=3,\ s=2,\ p=1\)
                    & \(32\times9\times16\) \\
        Conv + ReLU & \(32\to64,\ k=3,\ s=1,\ p=1\)
                    & \(64\times9\times16\) \\
        Adaptive average pooling & Output size \(2\times4\)
                                 & \(64\times2\times4\) \\
        Flatten & --- & \(512\) \\
        Linear + ReLU & \(512\to256\) & \(256\) \\
        \bottomrule
    \end{tabular}
\end{table}

\paragraph{Velocity estimator.}
The estimator concatenates \(h_t^v\) with the current
proprioceptive state, previous action, and proprioceptive history.
Its input dimension is \(256+64+29+372=721\).
An MLP with widths
\[
721\rightarrow512\rightarrow256\rightarrow3
\]
uses ELU after its first two affine layers and a linear output.
It predicts current root velocity in a yaw-aligned frame,
whose vertical axis remains world-up.
Behavioral latents, heading instructions, desired velocities,
and privileged simulator features are excluded from its input.

The estimator is supervised by same-timestep simulator velocity:
\begin{equation}
\mathcal L_{\mathrm{est}}
=
\frac{1}{3N}
\sum_{i=1}^{N}
\left\|
\widehat{\mathbf v}^{\mathrm{yaw}}_i
-
\mathbf v^{\mathrm{yaw,sim}}_i
\right\|_2^2.
\label{eq:app-estimator-loss}
\end{equation}
This loss has coefficient 1 and updates both the estimator
and the shared visual encoder.
The policy receives
\(\operatorname{sg}(\widehat{\mathbf v}^{\mathrm{yaw}})\),
where \(\operatorname{sg}\) denotes stop-gradient.
Policy gradients still update the visual encoder through
the direct visual-feature input.

\subsection{Actor Objective and Critic Coefficients}
\label{app:actor-objective}

For each ensemble, define the conservative estimate
\begin{equation}
Q_X^-=\min(Q_{X,1},Q_{X,2}),
\qquad
Q_{\mathrm{FB},k}=F_k(o,z,a)^\top z.
\end{equation}
Let \(\langle\cdot\rangle\) denote a minibatch average and
\(\langle f\rangle_m\) denote an average over samples with
validity mask \(m\).
We use the detached FB scale
\begin{equation}
\alpha
=
\operatorname{sg}
\left(
\left\langle |Q_{\mathrm{FB}}^-| \right\rangle
\right).
\end{equation}
The actor and estimator are optimized with
\begin{equation}
\begin{aligned}
\mathcal L_{\mathrm{actor}}
={}&
-\left\langle Q_{\mathrm{FB}}^- \right\rangle\\
&-\alpha\Big[
0.01\left\langle Q_D^- \right\rangle
+0.02\left\langle Q_A^- \right\rangle\\
&\hspace{2.5em}
+0.002\left\langle Q_H^- \right\rangle_{m_H}
+0.002\left\langle Q_V^- \right\rangle_{m_V}
\Big]
+\mathcal L_{\mathrm{est}}.
\end{aligned}
\label{eq:app-actor-objective}
\end{equation}

\begin{table}[t]
    \centering
    \small
    \caption{Coefficients in the actor objective.
    The four auxiliary value terms are additionally multiplied
    by the detached scale \(\alpha\).}
    \label{tab:actor-coefficients}
    \begin{tabular}{l l r}
        \toprule
        Value term & Role & Base coefficient \\
        \midrule
        \(Q_{\mathrm{FB}}\) & Behavioral intent & 1 \\
        \(Q_D\) & Conditional motion prior & 0.01 \\
        \(Q_A\) & Physical auxiliary objectives & 0.02 \\
        \(Q_H\) & Heading tracking & 0.002 \\
        \(Q_V\) & Horizontal-velocity tracking & 0.002 \\
        \bottomrule
    \end{tabular}
\end{table}

The coefficients in Table~\ref{tab:actor-coefficients} weight
critic values in the actor objective.
They are distinct from the instantaneous reward coefficients
used to train each critic.
The heading and velocity terms use their collected instruction
contexts and valid samples rather than relabeled instruction targets.

\paragraph{Conditional motion-prior reward.}
The discriminator is trained with binary cross-entropy,
using expert samples as positives and policy samples as negatives.
A gradient penalty with coefficient 10 is applied to the
discriminator logit on interpolated expert--policy inputs.
The reward for \(Q_D\) is
\begin{equation}
r_D(x,z)
=
\log p(x,z)-\log(1-p(x,z)),
\end{equation}
where \(p\) is the discriminator probability clipped to
\([10^{-6},1-10^{-6}]\).

\paragraph{Heading reward.}
On valid heading transitions, the reward is
\begin{equation}
r_{H,t}
=
-\left[
1-\cos\left(
\psi_{t+1}-\psi^\star_{t+1}
\right)
\right].
\label{eq:app-heading-reward}
\end{equation}
The executed and target headings are both evaluated at the
next state.
Invalid instruction transitions are masked from the corresponding
critic and actor objectives.

\paragraph{Horizontal-velocity reward.}
On valid velocity transitions, the reward is
\begin{equation}
r_{V,t}
=
-\mathcal H_1
\left(
\frac{
\left\|
\mathbf v^{\mathrm{sim}}_{t+1,xy}
-
\mathbf v^\star_{t,xy}
\right\|_2
}{
0.5\,\mathrm{m/s}
}
\right),
\label{eq:app-velocity-reward}
\end{equation}
where
\begin{equation}
\mathcal H_1(u)
=
\begin{cases}
\frac12 u^2, & 0\leq u\leq1,\\
u-\frac12, & u>1.
\end{cases}
\end{equation}
The reward compares post-action velocity with the current
velocity instruction that generated the action.
It applies Huber loss to the horizontal error-vector norm,
rather than independently to its components.
Valid zero-velocity commands remain supervised stopping targets.
Heading and velocity bootstrapping is interrupted at
instruction-context boundaries.

\subsection{Physical Auxiliary Rewards}
\label{app:physical-rewards}

The auxiliary critic \(Q_A\) learns from six physical costs.
Their raw weighted sum is
\begin{equation}
\begin{aligned}
r_A^{\mathrm{raw}}
={}&
-0.1c_{\mathrm{rate}}
-10c_{\mathrm{limit}}
-c_{\mathrm{impact}}\\
&-4c_{\mathrm{slip}}
-c_{\mathrm{ankle}}
-2c_{\mathrm{stumble}}.
\end{aligned}
\label{eq:app-aux-reward}
\end{equation}

\begin{table}[t]
    \centering
    \small
    \caption{Physical costs and their coefficients inside
    \(r_A^{\mathrm{raw}}\). These coefficients are separate
    from the actor coefficient \(0.02\) for \(Q_A\).}
    \label{tab:physical-reward-weights}
    \begin{tabular}{l r}
        \toprule
        Cost & Coefficient \\
        \midrule
        Action change & \(-0.1\) \\
        Joint-position limit violation & \(-10\) \\
        Body-impact severity & \(-1\) \\
        Loaded-foot tangential speed & \(-4\) \\
        Ankle-roll magnitude & \(-1\) \\
        Foot stumbling & \(-2\) \\
        \bottomrule
    \end{tabular}
\end{table}

\paragraph{Action and joint costs.}
Let \([x]_+=\max(x,0)\).
The action-change and joint-limit costs are
\begin{align}
c_{\mathrm{rate}}
&=
\sum_{j=1}^{29}
\left(
a^{\mathrm{env}}_{t,j}
-
a^{\mathrm{env}}_{t-1,j}
\right)^2,\\
c_{\mathrm{limit}}
&=
\sum_{j=1}^{29}
\left(
[q_j^- - q_j]_+
+
[q_j-q_j^+]_+
\right).
\end{align}
Here, \(a^{\mathrm{env}}\) denotes the rescaled environment
command before action delay.
The normalized policy action is scaled by 5 and clipped
to \([-5,5]\) for this command.
The soft joint limits \((q_j^-,q_j^+)\) retain 95\% of
the hard joint range around its midpoint.
The ankle cost is
\begin{equation}
c_{\mathrm{ankle}}
=
q_{\mathrm{left\ ankle\ roll}}^2
+
q_{\mathrm{right\ ankle\ roll}}^2.
\end{equation}

\paragraph{Contact-dependent costs.}
Let \(W\) be robot weight and
\(\widehat{\mathbf f}=\mathbf f/(\|\mathbf f\|_2+10^{-6})\)
the regularized contact-force direction.
The training slippage cost is
\begin{equation}
c_{\mathrm{slip}}
=
\sum_{f\in\{L,R\}}
\mathbf 1\!\left[
\|\mathbf f_f\|_2/W>0.05
\right]
\left\|
\mathbf v_f
-
(\mathbf v_f^\top\widehat{\mathbf f}_f)
\widehat{\mathbf f}_f
\right\|_2.
\end{equation}
It penalizes foot-link velocity tangent to the contact-force
direction under load.
This training cost differs from the displacement-based FS
evaluation metric.

For body impact, let \(\mathbf f_b^*\) be the contact-force
vector with maximum magnitude during the control interval,
and \(\mathbf v_b^{\mathrm{pre}}\) the pre-step body velocity.
Define
\begin{equation}
S_\beta(x)
=
\begin{cases}
x^2/(2\beta), & |x|<\beta,\\
|x|-\beta/2, & |x|\geq\beta.
\end{cases}
\end{equation}
The impact cost is
\begin{equation}
\begin{aligned}
c_{\mathrm{impact}}
=
\min\Bigg\{4,\;
\sum_{b\in\mathcal B}
&S_{0.25}
\left(
[\|\mathbf f_b^*\|_2/W-0.05]_+
\right)\\
&\times
\left[
1+
\min\left(
\frac{
[\mathbf v_b^{\mathrm{pre}\top}
\widehat{\mathbf f_b^*}]_+
}{
1\,\mathrm{m/s}
},
2
\right)^2
\right]
\Bigg\}.
\end{aligned}
\end{equation}
The monitored set \(\mathcal B\) contains the configured
pelvis, shoulder, and hip links.
The force direction follows the simulator's contact convention,
so positive projected pre-step velocity represents incoming motion.

The stumbling cost uses foot-collision force histories:
\begin{equation}
c_{\mathrm{stumble}}
=
\max_{g,h}
\min\left\{
1,\;
\frac{
[\|\mathbf f_{g,h}^{xy}\|_2-|f_{g,h}^{z}|-20\,\mathrm{N}]_+
}{
100\,\mathrm{N}
}
\right\},
\end{equation}
where \(g\) indexes foot-collision contact slots and \(h\)
indexes physics substeps within the control interval.

\paragraph{Dynamic reward normalization.}
The combined auxiliary reward is divided by a running
standard deviation:
\begin{equation}
\widetilde r_A
=
\frac{r_A^{\mathrm{raw}}}{\sigma_A}.
\end{equation}
The normalizer uses bias-corrected exponential moving averages
of the first and second moments, with decay \(0.99\) and
a variance floor of \(10^{-8}\).
The statistics are updated during training.
The output is scaled without subtracting the running mean,
and normalization is applied to the combined reward rather
than separately to each cost.
The auxiliary reward supplied to \(Q_A\) has no additional
control-timestep multiplier.

\subsection{Optimization Hyperparameters}
\label{app:optimization}

We use Adam optimization with the hyperparameters in
Table~\ref{tab:optimization}.
The forward and backward representations are trained with the
FB successor-measure objective and backward-embedding
orthonormality regularization.
No additional implicit-reward scalar \(Q\)-regression loss is enabled.

\begin{table}[t]
    \centering
    \small
    \caption{Optimization settings of the evaluated full model.}
    \label{tab:optimization}
    \begin{tabular}{l l}
        \toprule
        Parameter & Value \\
        \midrule
        Latent dimension & 256 \\
        Discount factor & 0.98 \\
        Minibatch size per GPU & 1,024 \\
        Effective minibatch size across eight GPUs & 8,192 \\
        Forward-network learning rate & \(3\times10^{-4}\) \\
        Backward-network learning rate & \(1\times10^{-5}\) \\
        Actor learning rate & \(3\times10^{-4}\) \\
        Scalar-critic learning rates & \(3\times10^{-4}\) \\
        Discriminator learning rate & \(5\times10^{-6}\) \\
        Backward orthonormality coefficient & 100 \\
        Discriminator gradient-penalty coefficient & 10 \\
        Velocity-estimation loss coefficient & 1 \\
        Action sampling standard deviation & 0.05 \\
        Action sampling noise clip & 0.3 \\
        Forward/backward target-update rate & 0.01 \\
        Scalar-critic target-update rate & 0.005 \\
        Physics frequency & 200 Hz \\
        Policy frequency & 50 Hz \\
        \bottomrule
    \end{tabular}
\end{table}

Target parameters are updated as
\begin{equation}
\bar\theta
\leftarrow
(1-\tau)\bar\theta+\tau\theta.
\end{equation}
The scalar-critic bootstrap targets use the minimum of their
two ensemble predictions.
The FB successor-measure bootstrap target instead uses the
mean of the two forward-network predictions.
For actor optimization, both FB values and scalar-critic
values use conservative minimum estimates.
\subsection{Terrain Configuration}
\label{app:terrains}

Training and evaluation use the 7 terrain families listed
in Table~\ref{tab:training-terrains}.
The terrain map contains 10 rows and 7 columns, with 1
column per terrain family.
Each tile measures \(5\times5\,\mathrm{m}\), giving a
\(50\times35\,\mathrm{m}\) main terrain grid.
Each non-flat tile has a \(0.5\,\mathrm{m}\)-wide flat border
on all four sides, leaving a \(4\times4\,\mathrm{m}\) interior.

\paragraph{Difficulty configuration.}
Rows are indexed by \(r\in\{0,\ldots,9\}\).
The normalized difficulty of the tile in row \(r\) and column
\(c\) is sampled as
\begin{equation}
d_{r,c}=\frac{r+u_{r,c}}{10},
\qquad
u_{r,c}\sim\mathcal{U}(0,1).
\label{eq:app-terrain-difficulty}
\end{equation}
Thus, each row specifies a difficulty interval rather than
one identical difficulty value for every terrain family.
Terrain generation uses a fixed seed of 0, and all evaluated
methods share the same generated evaluation map.
The parameters below describe the configured generation
ranges; the finite set of generated tiles need not attain
their endpoints.
For the flat terrain family, all ten rows remain physically flat.

\begin{table}[H]
    \centering
    \small
    \renewcommand{\arraystretch}{1.18}
    \caption{Terrain families and difficulty-dependent geometry.
    IDs follow the terrain generator's column order.
    Here \(d\) denotes normalized tile difficulty.
    Stair traversal directions are defined for motion from
    the central platform toward the tile boundary.}
    \label{tab:training-terrains}
    \begin{tabular}{@{}c p{0.25\linewidth} p{0.63\linewidth}@{}}
        \toprule
        \textbf{ID} & \textbf{Terrain family}
        & \textbf{Geometry parameters} \\
        \midrule
        0 & Flat terrain
        & Planar surface; geometry is independent of difficulty. \\

        1 & Perlin-noise rough terrain
        & Target height range \(0\) to
          \(H_{\mathrm{rough}}(d)=0.02+0.04d\,\mathrm{m}\)
          before border resampling. \\

        2 & Pyramid stairs
            (outward descent)
        & Riser height \(h(d)=0.10+0.10d\,\mathrm{m}\);
          tread depth \(0.30\,\mathrm{m}\);
          central platform \(1.6\times1.6\,\mathrm{m}\). \\

        3 & Inverted pyramid stairs
            (outward ascent)
        & Same riser heights, tread depth, and central-platform
          dimensions as the pyramid stairs, with inverted
          elevation. \\

        4 & Low platforms
        & Grid-cell width \(0.80\,\mathrm{m}\);
          cell heights sampled from
          \([-H_{\mathrm{plat}}(d),H_{\mathrm{plat}}(d)]\),
          where \(H_{\mathrm{plat}}(d)=0.03+0.07d\,\mathrm{m}\),
          before height quantization. \\

        5 & Inverted pyramid slope
        & Configured slope parameter \(s(d)=0.20d\);
          nominal angle envelope
          \(0\) to \(\arctan(0.20)\approx11.31^\circ\). \\

        6 & Box-obstacle terrain
        & Height bound
          \(H_{\mathrm{box}}(d)=0.05+0.10d\,\mathrm{m}\);
          sampled rectangle side lengths
          \(\{0.4,0.6,0.8,1.0\}\,\mathrm{m}\);
          50 placement attempts per tile. \\
        \bottomrule
    \end{tabular}
\end{table}

\paragraph{Terrain generation details.}
Rough terrain uses two Perlin-noise octaves, persistence
\(0.25\), lacunarity \(2\), and a noise-scale parameter of \(20\).
The sampled noise is normalized to \([0,1]\) before multiplication
by \(H_{\mathrm{rough}}(d)\).
Subsequent interpolation can reduce the realized height range.

Both stair families contain four concentric tread bands and a
central platform, producing five vertical transitions between
the tile border and the center.
Their central elevations are \(+5h(d)\) and \(-5h(d)\),
respectively.
The configured minimum central-platform width is
\(1.2\,\mathrm{m}\); fitting an integer number of treads into
the tile yields an actual width of \(1.6\,\mathrm{m}\).

Low-platform terrain uses a \(5\times5\) grid within the tile
interior.
Sampled cell heights are rounded to increments of
\(0.025\,\mathrm{m}\).
Adjacent cells with equal quantized heights may be merged,
with each merged rectangle spanning at most three cells
along either axis.
A separate \(1.2\times1.2\,\mathrm{m}\) central platform has
surface height \(H_{\mathrm{plat}}(d)\).

Box-obstacle terrain includes both positive and negative
height offsets, using full- and half-height levels derived
from \(H_{\mathrm{box}}(d)\) with \(0.005\,\mathrm{m}\)
vertical discretization.
Rectangle dimensions are sampled independently along the
two horizontal axes.
Placements may overlap and overwrite earlier placements;
therefore, 50 placement attempts do not imply 50 distinct
non-overlapping obstacles.
The slope and box generators use a nominal central-platform
or clearance-width parameter of \(1.2\,\mathrm{m}\).

The rough, slope, and box heightfields use a nominal
\(0.2\,\mathrm{m}\) source sampling scale.
They are subsequently resampled to \(31\times31\) vertices
per tile to enforce the flat-border geometry.
Consequently, final surface extrema and local slopes can
differ from their nominal generation parameters.

\paragraph{Map boundaries and evaluation labels.}
The main grid is surrounded by a \(10\,\mathrm{m}\)-wide
region of additional rough and box terrain, followed by a
\(2\,\mathrm{m}\)-wide flat outer band.
The additional terrain interiors are elevated by
\(0.05\,\mathrm{m}\), while their tile borders remain flat.
Perimeter walls are \(5\,\mathrm{m}\) high and
\(0.05\,\mathrm{m}\) thick.

Evaluation covers all seven initial terrain types and all
ten terrain rows.
Terrain labels describe the initial tile of each rollout.
The robot may subsequently enter neighboring tiles, so these
labels do not necessarily describe the entire executed trajectory.

\subsection{Reference Motions and Evaluation Coverage}
\label{app:motion-coverage}

All evaluations use fixed model weights without evaluation-time
optimization. Reference motions and executed states are compared at
50~Hz. References are placed in the evaluation scene using the initial
alignment, and the reference clock advances throughout each rollout.
Errors are evaluated against the corresponding reference time without
post-hoc temporal alignment.

\paragraph{LaFAN.}
The LaFAN evaluation uses 40 source motion sequences.
For the four continuous metrics, the three-method comparison covers
902 segments, including short terminal segments, whereas the ablation
evaluation covers 862 complete segments.
Both recovery evaluations use the same 862 complete 10-second segments.
A segment is an evaluation window and need not contain a complete skill,
particularly when a falling-and-getting-up sequence spans multiple segments.

\paragraph{SOMA.}
The continuous-metric evaluation uses all 994 selected recordings,
evaluated over their full durations.
The original 120~Hz references are resampled to 50~Hz.
For recovery evaluation, 180 recordings contain at least one complete
10-second window.
These recordings yield 281 consecutive, non-overlapping 10-second
segments after incomplete terminal windows are discarded.
Thus, the continuous metrics and recovery rates use different
evaluation sets.

The SOMA recordings provide an additional motion set beyond the
40 LaFAN sequences used in the reported training configuration.

\paragraph{Released PMT references.}
The additional PMT-reference evaluation uses 297 segments from
99 released files.
These files contain only one unique joint-position sequence and
therefore do not represent 99 independent motion skills.
For recovery evaluation, the third segment starts at zero-based frame
999 rather than frame 1000.
This produces 501 reference states spanning 500 control intervals,
or exactly 10 seconds, without extrapolation.
The third segment consequently overlaps the preceding segment by
one control interval.

\paragraph{Evaluation counts.}
Each continuous-metric condition is evaluated with five paired trials.
Recovery uses one trial per condition at each of four force magnitudes.
All conditions cover seven initial terrain types and ten terrain rows.
Table~\ref{tab:evaluation-coverage} reports the scheduled counts per model.

\begin{table}[t]
    \centering
    \small
    \caption{Main evaluation coverage per model.
    Continuous metrics use five trials per motion--terrain--row
    condition. Recovery uses one trial at each of four force magnitudes.}
    \label{tab:evaluation-coverage}
    \begin{tabular}{l r r r r}
        \toprule
        Evaluation & \multicolumn{2}{c}{Continuous metrics}
                   & \multicolumn{2}{c}{Recovery} \\
        \cmidrule(lr){2-3}\cmidrule(lr){4-5}
        & Motions/segments & Episodes & Segments & Episodes \\
        \midrule
        LaFAN comparison & 902 & 315,700 & 862 & 241,360 \\
        LaFAN ablation   & 862 & 301,700 & 862 & 241,360 \\
        SOMA             & 994 & 347,900 & 281 & 78,680 \\
        PMT references   & 297 & 103,950 & 297 & 83,160 \\
        \bottomrule
    \end{tabular}
\end{table}

\subsection{Continuous Tracking Metrics}
\label{app:continuous-metrics}

Let \(T_i\) denote the number of recorded post-action states in
episode \(i\).
The following metrics first produce one scalar per episode.
The reference and executed quantities are expressed using the same
joint ordering and scene-coordinate convention.

\paragraph{Joint-angle error.}
For the 29 actuated joints, we compute
\begin{equation}
E_{\mathrm{jpos},i}
=
\frac{1}{29T_i}
\sum_{t=1}^{T_i}
\sum_{j=1}^{29}
\left|
q_{i,t,j}-q^{\mathrm{ref}}_{i,t,j}
\right|.
\label{eq:app-joint-error}
\end{equation}
The unit is radians.
Despite the notation \(E_{\mathrm{jpos}}\), this metric measures
joint-angle MAE, not Cartesian joint-position error.

\paragraph{Root horizontal-velocity error.}
We use simulator ground-truth root velocity rather than the
policy's velocity estimate:
\begin{equation}
e^v_{i,t}
=
\left\|
\mathbf v^{\mathrm{sim}}_{i,t,xy}
-
\mathbf v^{\mathrm{ref}}_{i,t,xy}
\right\|_2,
\qquad
E_{\mathrm{vel},i}^{\mathrm{root}}
=
\frac{1}{T_i}\sum_{t=1}^{T_i}e^v_{i,t}.
\label{eq:app-velocity-error}
\end{equation}
Both vectors are expressed in world horizontal coordinates.
The unit is \(\mathrm{m/s}\).
The metric compares velocity vectors, not their scalar speeds.

\paragraph{Root-heading error.}
Heading is obtained from the horizontal projection of the root's
forward axis.
Let \(\psi_{i,t}\) and \(\psi^{\mathrm{ref}}_{i,t}\) denote the
executed and reference headings.
The shortest absolute angular difference is
\begin{equation}
e^h_{i,t}
=
\frac{180}{\pi}
\left|
\operatorname{atan2}
\left(
\sin\Delta\psi_{i,t},
\cos\Delta\psi_{i,t}
\right)
\right|,
\qquad
\Delta\psi_{i,t}
=
\psi_{i,t}-\psi^{\mathrm{ref}}_{i,t}.
\label{eq:app-heading-error}
\end{equation}
Thus, \(e^h_{i,t}\in[0,180]\) degrees.
For the continuous metric, a frame is valid only when both
forward-axis horizontal projections have norm greater than \(10^{-6}\).
Writing \(\mathcal V_i\) for this set,
\begin{equation}
E_{\mathrm{head},i}^{\mathrm{root}}
=
\frac{1}{|\mathcal V_i|}
\sum_{t\in\mathcal V_i}e^h_{i,t}.
\end{equation}
The valid-frame fraction is recorded separately.
An episode with no valid heading frames has an undefined heading
metric rather than zero error.
The same validity rule is applied to every method, although the
resulting masks may differ because executed orientations differ.

\paragraph{Foot sliding.}
FS measures horizontal displacement of the left and right
ankle-roll link origins during consecutive contact frames.
Let \(\mathbf p^{xy}_{i,f,t}\) be the world horizontal position
of foot-link origin \(f\in\{L,R\}\), measured in meters.
Contact is detected using the recorded vertical contact-force component:
\begin{equation}
c_{i,f,t}
=
\mathbf 1\!\left[|F^z_{i,f,t}|>1\,\mathrm{N}\right].
\end{equation}
We compute
\begin{equation}
\mathrm{FS}_i
=
\frac{100}{2(T_i-1)}
\sum_{f\in\{L,R\}}
\sum_{t=2}^{T_i}
c_{i,f,t-1}c_{i,f,t}
\left\|
\mathbf p^{xy}_{i,f,t}
-
\mathbf p^{xy}_{i,f,t-1}
\right\|_2.
\label{eq:app-fs}
\end{equation}
The factor 100 converts meters to centimeters.
FS is reported in cm and represents contact-masked
horizontal displacement averaged over both feet and all
recorded control intervals at 50~Hz.
The denominator includes non-contact intervals.
Displacement magnitudes are summed before averaging,
rather than using the net displacement between endpoints.

FS is a foot-link sliding proxy rather than a direct measurement
of physical contact-point slip.
Foot rotation around a stationary contact can move the measured
link origin, and the score also depends on contact occupancy.
If either foot has no valid consecutive-contact interval,
the episode-level FS is undefined and is excluded from its
metric-specific mean, with the valid episode count retained.
Undefined values are not replaced by zero.

\subsection{Perturbation and Recovery Protocol}
\label{app:recovery}

\paragraph{Perturbations.}
Each recovery rollout spans 10 seconds.
A horizontal force is applied to the pelvis from
\(t=2.0\) to \(t=2.2\) seconds while the reference clock
continues to advance.
The force magnitude is selected from
\(\{50,100,150,200\}\,\mathrm{N}\).
For direction \(\theta\), the world-frame force is
\begin{equation}
\mathbf F
=
F
\begin{bmatrix}
\cos\theta & \sin\theta & 0
\end{bmatrix}^{\top}.
\end{equation}
Directions are deterministically generated from the trial seed
and condition identifiers.
Matched conditions share the same force direction across methods
and force magnitudes.
One trial is evaluated for each motion--terrain--row--force condition.

\paragraph{Recovery criterion.}
At each recorded state, define
\begin{equation}
b_i(t)
=
\mathbf 1\!\left[
e^h_i(t)<30^\circ
\;\land\;
e^v_i(t)\leq0.5\,\mathrm{m/s}
\;\land\;
\text{both errors are finite and valid}
\right].
\label{eq:app-recovery-tolerance}
\end{equation}
The velocity threshold applies to the norm of the horizontal
error vector at each time step.
It is not applied separately to its components or to an
error accumulated over time.
Undefined heading frames do not satisfy the recovery criterion.

Let \(t_{\mathrm{off}}=2.2\,\mathrm{s}\) and
\(t_{\mathrm{end}}=10\,\mathrm{s}\).
For a recovery deadline \(w\in\{3,5\}\,\mathrm{s}\), a trial succeeds
if the full observation window is completed without reset and
there exists an entry time \(\tau\) satisfying
\begin{equation}
\begin{aligned}
&t_{\mathrm{off}}\leq\tau\leq t_{\mathrm{off}}+w,\\
&t_{\mathrm{end}}-\tau\geq1\,\mathrm{s},\\
&b_i(t)=1
\quad\text{at every recorded state from }\tau
\text{ through }t_{\mathrm{end}}.
\end{aligned}
\label{eq:app-recovery-success}
\end{equation}
Therefore, a temporary return to tolerance followed by another
violation is insufficient.
Trials remaining within tolerance throughout the post-perturbation
period also count as successful.

Writing \(S_i^{(w)}\) for this binary outcome, the main-table rate is
\begin{equation}
\mathrm{RR}@w
=
100\,
\frac{1}{N_{\mathrm{scheduled}}}
\sum_{i=1}^{N_{\mathrm{scheduled}}}S_i^{(w)}.
\label{eq:app-recovery-rate}
\end{equation}
All scheduled trials remain in the denominator.
Trials that reset or fail to complete the observation window
cannot satisfy the success criterion.
In the main text and main tables, RR denotes
\(\mathrm{RR}@5\mathrm{s}\).
We retain explicit deadline labels when reporting both
\(\mathrm{RR}@3\mathrm{s}\) and \(\mathrm{RR}@5\mathrm{s}\)
in supplementary results.

Tracking violations during the one-second interval before
force application are recorded separately.
They do not automatically invalidate a subsequent recovery.
This metric measures recovery or maintenance of heading and
horizontal-velocity tracking.
It does not require returning to the original horizontal path,
satisfying a joint-angle threshold, or recovering an upright
whole-body pose.
Consequently, it should not be interpreted as a general
fall-recovery success rate.

\subsection{Aggregation and Incomplete Rollouts}
\label{app:aggregation}

For each continuous metric, let \(E_i\) be the episode-level
value and \(N\) its number of valid episodes.
We report
\begin{equation}
\mu=\frac{1}{N}\sum_{i=1}^{N}E_i,
\qquad
s=
\sqrt{
\frac{1}{N-1}
\sum_{i=1}^{N}(E_i-\mu)^2
}.
\label{eq:app-mean-std}
\end{equation}
Episodes are equally weighted across segments or recordings,
initial terrains, terrain rows, and evaluation repeats.
The reported standard deviation describes variation among
episode-level means.
It is neither a within-episode temporal standard deviation
nor a standard deviation across five independently trained models.

The main tables do not use source-motion macro-averaging.
Consequently, source motions contributing more segments receive
greater weight in segment-based evaluations.
For full-recording SOMA evaluation, each scheduled recording
condition receives equal weight regardless of recording duration.
Metric-specific validity counts are retained because undefined
heading or FS values can change the effective sample size.

Ordinary tracking failures do not trigger automatic resets when
simulation can continue.
Continuous metrics therefore include subsequent execution rather
than only the successful portion of a rollout.
In the SOMA comparison, 8 PMT episodes and 219 UFO episodes ended
early at the global environment boundary.
Their recorded-time means remain included.
These episodes have shorter observation windows, which limits
their direct comparability with full-duration episodes.
Program errors, non-finite states, and policy tracking failures
are recorded separately.

\subsection{Category-Balanced SOMA Recovery Supplement}
\label{app:balanced-soma}

We additionally evaluate a category-balanced SOMA subset to
examine recovery under a different motion composition.
The subset contains 30 recordings from each of six categories:
walking, running, dancing, jumping, low-pose motions, and sports.
The 180 recordings comprise 161 motion-name groups and yield
255 complete 10-second segments.
Incomplete terminal windows are discarded.
This evaluation uses the same perturbation and recovery criteria
as Appendix~\ref{app:recovery}.

For this supplement, outcomes are first averaged over conditions
and segments within each recording.
Recording-level rates are then averaged within each category,
followed by an equal-weight average over the six categories.
This differs from the episode-weighted aggregation of the
281-segment SOMA recovery set used in the main tables.

\begin{table}[t]
    \centering
    \small
    \caption{Category-balanced SOMA recovery results.
    Rates are recording- and category-macro averages.
    Each condition uses one trial per force magnitude.
    Pre-push violation denotes failure to remain within the
    tracking tolerances throughout the one-second pre-push interval.}
    \label{tab:balanced-soma-rr}
    \begin{tabular}{l r r r}
        \toprule
        Method
        & RR@3s (\%)
        & RR@5s (\%)
        & Pre-push violation (\%) \\
        \midrule
        \multicolumn{4}{l}{\textit{Three-method comparison}} \\
        UFO      & 22.50 & 28.38 & 31.70 \\
        PMT      & 0.87  & 1.07  & 98.98 \\
        \ours{}  & 56.99 & 69.50 & 20.30 \\
        \midrule
        \multicolumn{4}{l}{\textit{Ablation}} \\
        w/o Heading Instr.  & 35.15 & 45.44 & 21.52 \\
        w/o Velocity Instr. & 37.03 & 52.31 & 32.09 \\
        w/o Perception      & 54.64 & 66.04 & 21.54 \\
        \ours{}             & 57.03 & 69.56 & 20.31 \\
        \bottomrule
    \end{tabular}
\end{table}

The balanced results are reported separately and do not replace
the original SOMA main-table values.
In particular, PMT's high pre-push violation rate indicates that
its recovery score also reflects tracking difficulty already
present before perturbation.
Differences in the aggregate rates therefore cannot be attributed
solely to the applied forces.

\subsection{Real-World Depth Acquisition and Preprocessing}
\label{app:realworld-depth}

The camera and depth-processing settings recorded for the final
real-world deployment are summarized in
Table~\ref{tab:realworld-depth}.
Automatic exposure was disabled, with exposure time fixed at
\(1600\,\mu\mathrm{s}\).
The depth stream was configured at
\(480\times270\) pixels and \(60\,\mathrm{Hz}\).
The depth encoder ran on the CPU with a configured update
frequency of \(50\,\mathrm{Hz}\).
The camera acquisition rate and encoder update rate are
distinct settings; the latter is configurable.

\begin{table}[H]
    \centering
    \small
    \caption{Recorded camera and depth-processing settings
    for the final real-world deployment.
    Gain and infrared emitter power are reported as device
    setting values.}
    \label{tab:realworld-depth}
    \begin{tabular}{l l}
        \toprule
        \textbf{Parameter} & \textbf{Setting} \\
        \midrule
        Automatic exposure          & Disabled \\
        Exposure time               & \(1600\,\mu\mathrm{s}\) \\
        Gain                        & 16 \\
        Infrared emitter power      & 240 \\
        Depth-stream resolution     & \(480\times270\) pixels \\
        Depth-stream frame rate     & \(60\,\mathrm{Hz}\) \\
        Depth-encoder device        & CPU \\
        Configured encoder rate     & \(50\,\mathrm{Hz}\) \\
        Vertical image flip         & Disabled \\
        Spatial filter              & Disabled \\
        Temporal filter             & Disabled \\
        Hole-filling filter         & Disabled \\
        Gaussian smoothing          & Enabled, \(3\times3\) kernel \\
        Encoder input per frame     & \(18\times32\) pixels \\
        Maximum depth               & \(2.5\,\mathrm{m}\) \\
        Invalid-depth replacement   & \(2.5\,\mathrm{m}\) \\
        Normalized input range      & \([0,1]\) \\
        \bottomrule
    \end{tabular}
\end{table}

\paragraph{Depth preprocessing.}
Depth measurements were mapped to the training ray layout
using nearest-neighbor sampling and then cropped to
\(18\times32\) pixels per frame.
The maximum depth was set to \(2.5\,\mathrm{m}\), and invalid
measurements were replaced with this maximum value,
consistent with the training preprocessing convention.
Depth inputs were normalized to \([0,1]\) before encoding.
The final deployment enabled \(3\times3\) Gaussian smoothing,
with spatial, temporal, and hole-filling filters disabled.

\end{document}